\pdfoutput=1

\documentclass[11pt]{article}

\usepackage[]{acl}

\usepackage{times}
\usepackage{latexsym}

\usepackage{graphicx}
\usepackage{algorithm}
\usepackage{algorithmic}
\usepackage{amsmath,amsthm,mathtools}
\usepackage{booktabs}
\usepackage{multirow}
\usepackage{xcolor}
\usepackage{amsmath,amsthm,amssymb}

\usepackage[T1]{fontenc}

\usepackage[utf8]{inputenc}

\usepackage{microtype}

\title{Topic-Aware Response Generation in Task-Oriented Dialogue with Unstructured Knowledge Access}

\author{Yue Feng$^\dagger$ \thanks{\;\;The work was done when the first author was an intern at
Huawei Noah’s Ark Lab.} \quad Gerasimos Lampouras$^\ddagger$ \quad Ignacio Iacobacci$^\ddagger$\\
  $^\dagger$University College London, London, UK\\
  $^\ddagger$Huawei Noah’s Ark Lab, London, UK\\
  $^\dagger$ {\texttt{yue.feng.20@ucl.ac.uk}} \\
  $^\ddagger$ {\texttt{\{gerasimos.lampouras, ignacio.iacobacci\}@huawei.com}}\\}

\begin{document}
\maketitle
\begin{abstract}
To alleviate the problem of structured databases' limited coverage, recent task-oriented dialogue systems incorporate external unstructured knowledge to guide the generation of system responses. 
However, these usually use word or sentence level similarities to detect the relevant knowledge context, which only partially capture the topical level relevance. 
In this paper, we examine how to better integrate topical information in knowledge grounded task-oriented dialogue and propose ``Topic-Aware Response Generation'' (TARG), an end-to-end response generation model. TARG incorporates multiple topic-aware attention mechanisms to derive the importance weighting scheme over dialogue utterances and external knowledge sources towards a better understanding of the dialogue history. 
Experimental results indicate that TARG achieves state-of-the-art performance in knowledge selection and response generation, outperforming previous state-of-the-art by 3.2, 3.6, and 4.2 points in EM, F1 and BLEU-4 respectively on Doc2Dial, and performing comparably with previous work on DSTC9; both being knowledge-grounded task-oriented dialogue datasets.
\end{abstract}

\section{Introduction}
Task-oriented (or goal-oriented) dialogue systems aim to accomplish a particular task (e.g. book a table, provide information) through natural language conversation with a user. The system's available actions are often described by a pre-defined domain-specific schema while relevant knowledge is retrieved from stuctured databases or APIs \citep{feng2022dynamic,rastogi2020towards}. As such, task-oriented dialogue systems are often limited on which actions can be taken and what information can be retrieved~\citep{kim2020beyond}. To relax these restrictions, some dialogue systems (also referred to as goal-oriented chatbots) adopt open-domain language that is by definition unconstrained by pre-defined actions \citep{feng2020doc2dial}, and dynamically extract any required knowledge from in-domain unstructured collections in the form of entity descriptions, FAQs, and documents. Access to external knowledge sources has also been shown to help dialogue systems generate more specific and informative responses, which helps with the ``common response'' problem~\citep{zhang2018reinforcing, ren2020thinking, feng2021multi, feng2022multi, shi2022learning}.

\begin{figure}[!t]
\centering
\includegraphics[width=0.48\textwidth]{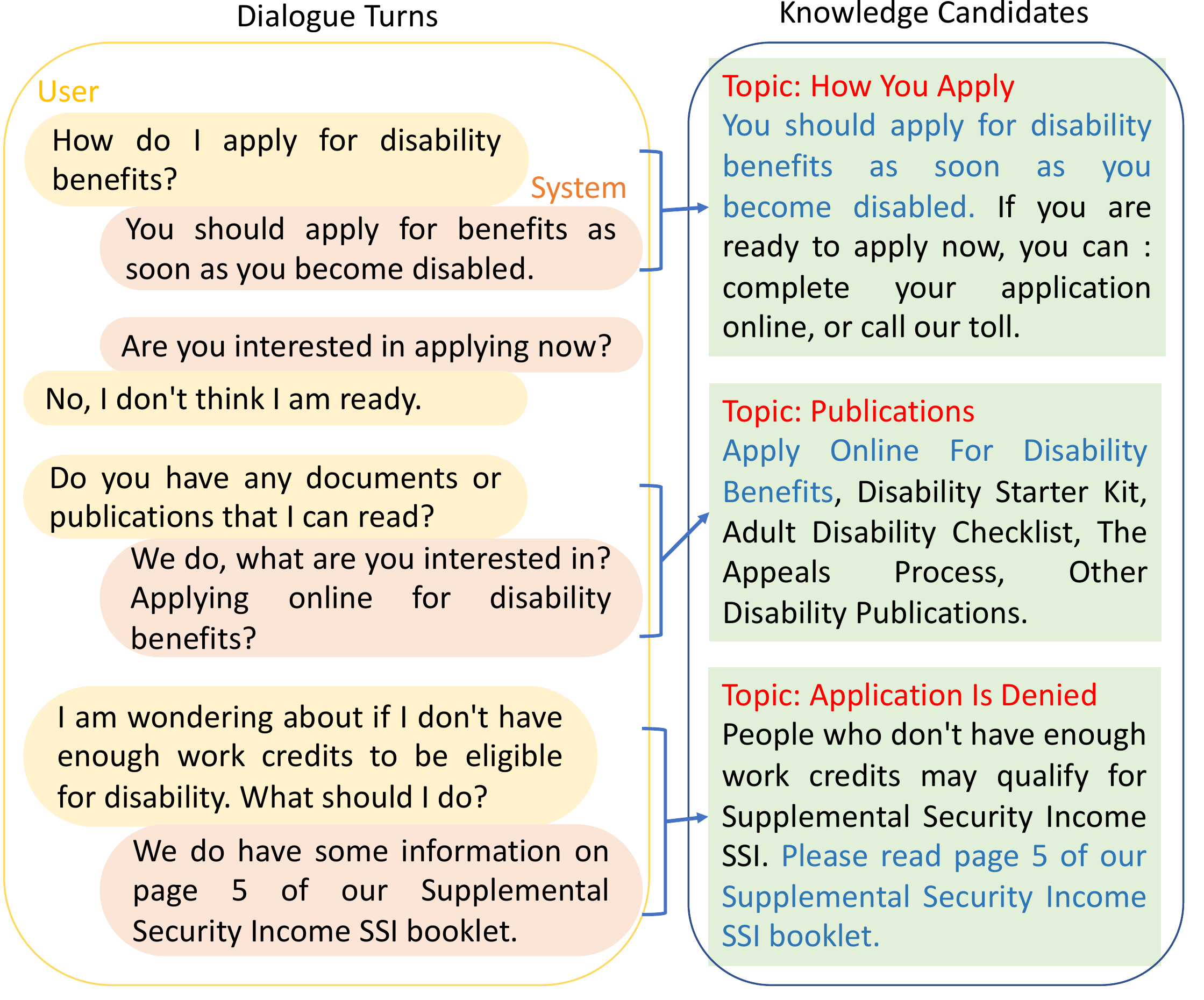}
\caption{An example of knowledge-grounded dialogue.}
\label{fig:dialogue_example}
\end{figure}

Figure~\ref{fig:dialogue_example} shows an example of a task-oriented dialogue that exploits external unstructured knowledge sources.
Given a history of previous dialogue turns, with each turn consisting of one user and system utterance, and access to in-domain unstructured knowledge sources (either a document collection or a set of candidate facts), the dialogue system needs to generate an appropriate system response for the current turn. Recent research~\citep{zhang2018reinforcing, ren2020thinking} tackles the task by decomposing it into two sub-tasks: to initially determine the relevant knowledge (if any) that needs to be extracted/selected from external resources, and to subsequently generate the response based on the selected knowledge and the dialogue history. 

When retrieving knowledge from unstructured sources, different sources may need to be accessed in different dialogue turns; this is to be expected in most conversation scenarios. In the example of Figure~\ref{fig:dialogue_example}, the first turn is grounded on the first knowledge candidate, and subsequent turns are grounded on later candidates. If we consider that each knowledge source belongs to a different topic or domain  (e.g. ``how you apply'', ``publications'', ``application is denied'' in our example), we can observe that as the knowledge selection shifts across sources during the course of the dialogue, a corresponding shift occurs between topics. 
Previous work has not actively exploited this, but we posit that attending the topic shifts in the dialogue history can provide signals that help distinguish relevant from irrelevant sources for knowledge selection, and that such topical information can help the model derive an importance weighting scheme over the dialogue history for better response generation. 


In this paper, we model topic shifts in selected knowledge sources to improve topic-aware knowledge selection and response generation in task-oriented dialogue, and propose “Topic-Aware Response Generation” (TARG), an end-to-end model for knowledge selection and response generation. Our approach incorporates multiple topic-aware attention mechanisms to derive the importance weighting scheme over previous utterances and knowledge sources, aiming for a better understanding of the dialogue history. In addition, TARG is built on top of recent breakthroughs in language representation learning by finetuning on the pre-trained language model BART \cite{lewis2020bart}. 

We conduct extensive experiments with two task-oriented dialogue datasets, namely Doc2Dial~\citep{feng2020doc2dial} and DSTC9~\citep{gunasekara2020overview}. Our results indicate that TARG\footnote{The code is available at \url{https://github.com/huawei-noah/noah-research/tree/master/NLP/TARG}.} is able to accurately select the appropriate knowledge source, and as a result generate more relevant and fluent responses, outperforming previous state-of-the-art by 3.2, 3.6, and 4.2 points in EM, F1 and BLEU-4 respectively on Doc2Dial, and performing comparably with previous work on DSTC9. Furthermore, we present an ablation study and a case study accompanied by analysis of the learned attention mechanisms.





\begin{figure*}[!th]
\centering
\includegraphics[width=0.99\textwidth]{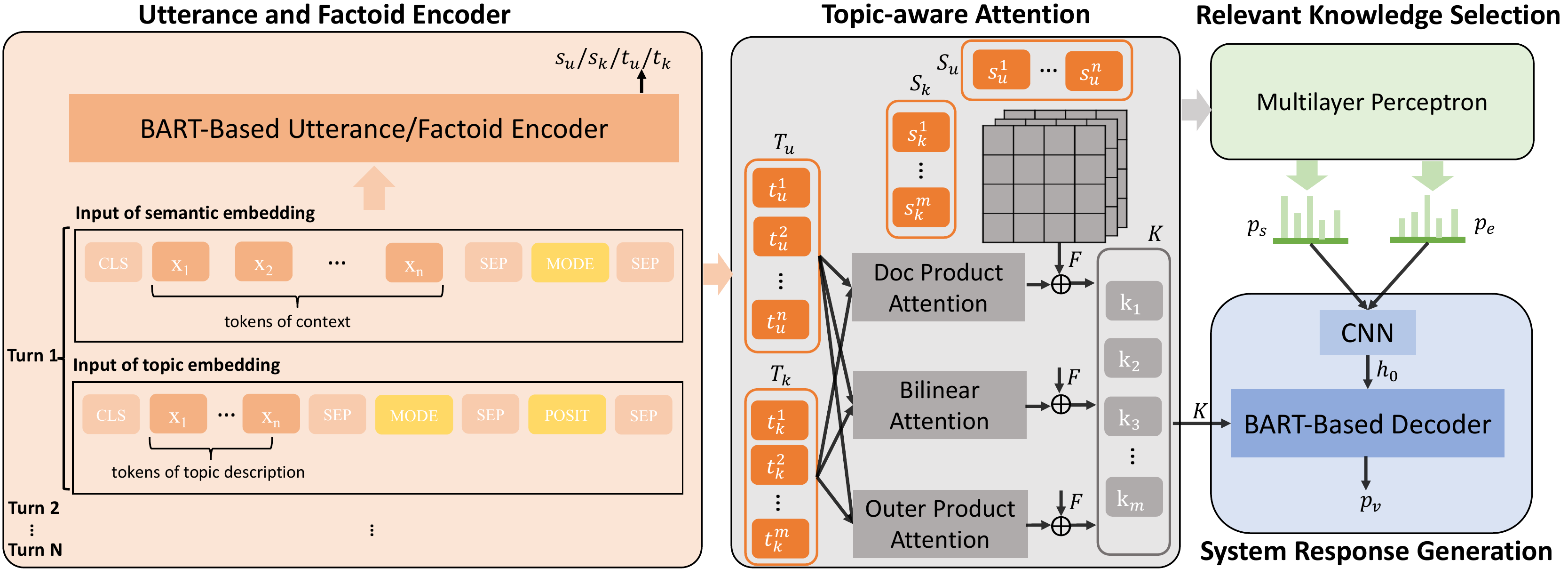}
\caption{Overview of Topic-Aware Response Generation (TARG). 
}
\label{fig:framework}
\end{figure*}

\section{Related Work}

As we briefly mentioned in the introduction, the majority of previous work decomposed knowledge-grounded dialogue generation into two sub-tasks: knowledge selection and response generation.

To determine the relevant candidate for knowledge selection, the use of keyword matching~\citep{ghazvininejad2018knowledge}, information retrieval~\citep{young2018augmenting} and entity diffusion~\citep{liu2018knowledge} methods have been proposed. More specifically, keyword matching methods~\citep{bordes2016learning} focus on calculating a weight for each 
keyword in the knowledge candidate and then determine their relevance based on the weighted sum of the keywords' representations. On the other hand, some information retrieval techniques compute traditional \textit{tf-idf} scores to detect the knowledge candidate in the most relevant document to the user’s query~\citep{song2018ensemble, dinan2018wizard}, while others leverage the power of neural networks to learn a candidate ranking function directly through an end-to-end learning process~\citep{yan2018coupled, zhao2019document, gu2019dually, gu2020filtering}. Another approach uses entity diffusion networks ~\citep{wang2020improving} that perform fact matching and knowledge diffusion to ground both knowledge candidates and dialogues.


For response generation, the related work 
has adapted both response retrieval and language generation approaches. Specifically for response retrieval, deep interaction networks~\citep{sun2020history} have been employed to learn better-suited representations to ground candidate responses against external knowledge, while language generation approaches have been adapted to attend to ground knowledge during inference~\citep{peng2020soloist}, with some further employing copy mechanisms over both dialogue context and external knowledge~\citep{yavuz2019deepcopy}, or leveraging a reading comprehension model to similarly extract relevant spans ~\citep{qin2019conversing,wu2021controllable}.


Recently, pre-trained language models such as BERT~\citep{devlin-etal-2019-bert} or RoBERTa~\citep{liu2019roberta}, which have demonstrated significant improvements on numerous natural language processing tasks, have also been applied to improve model the semantic representation in knowledge selection and response generation \citep{zhao2020knowledge,li2020zero, feng2020doc2dial, feng2021sequence, ye2022assist}. Alternatively, other approaches combine the generative capability of auto-regressive decoders such as GPT-2~\citep{budzianowski2019hello} or T5~\citep{2020t5}, to better generate the system response.

Broader dialogue research has explored the topic-aware signal present in the dialogue history, but such work did not consider external knowledge nor its topics. Briefly, \citet{Xing2017topic} proposed a topic-aware seq-to-seq approach for open-domain dialogue that attends over LDA topics inferred from the dialogue history, while \citet{zhang2020modeling} calculates the relevance between topic distributions of the dialogue history and the immediate context and attends over them to generate the next system response. In retrieval-based dialogue systems, \citet{xu2021topic} performs topic-aware segmentation of the context to better inform dialogue modeling.

We briefly discuss more recent work in our experiments section, as we compare it against our approach. To the best of our knowledge no other work has explicitly modelled the topic shifts in both dialogue history and external knowledge to inform knowledge selection and response generation in knowledge-ground task-oriented dialogue systems.

\section{Our Approach}
As we mentioned in the introduction, our proposed approach (TARG) exploits topic-aware mechanisms to derive an importance weighting scheme over different utterances in the dialogue history, with the goal to better inform knowledge selection and response generation. For a brief overview of TARG, please consult Figure~\ref{fig:framework}. The input in our task consists of the dialogue history of previous user and system utterances, and a set of external knowledge candidates (hereafter referred to as factoids for brevity). The goal is to generate the next system utterance in the dialogue, which may or may not be grounded in one of the factoids; some of the dialogue history utterances may also be grounded on factoids but not necessarily all of them are. 

Briefly, to generate the next turn's system utterance, TARG initially generates BART-based representations for every previous user and system utterance in the dialogue history, for every available factoid, and for both utterances' and factoids' corresponding topics. For each utterance / factoid pair, TARG extracts matching features by calculating feature interaction over their encoded representations. TARG subsequently weights the matching features by topic-aware attention mechanisms, and aggregates them in a tensor. Finally, a knowledge selection layer outputs a relevance score over factoids, and the decoder generates the system utterance based on the most relevant factoid's encoding. 

\subsection{Utterance and Factoid Encoder}
We use a BART encoder to generate representations for every utterance in the dialogue history (up to a maximum history length) and factoid in external knowledge. We similarly, but separately, generate representations for their corresponding topics. Our work assumes that the corresponding topic of factoids can be derived in some way from the available data, e.g. the topic can be interpreted as the title of the factoid's originating document or its annotated domain. While we do not explore the possibility in this paper, the topic could also potentially be inferred using topic modelling techniques. The topic of each utterances is considered the same as that of their corresponding factoids (if any). Since not all dialogue turns are necessarily grounded in external knowledge, in absence of a corresponding factoid, the topic is set to a generic ``non-relevant'' pseudo-topic. This process results in the semantics and topic of every utterance or factoid being represented explicitly by separate embeddings.


Specifically, in order to generate the semantic embeddings $s_u$ and $s_k$ of every utterance and factoid respectively, the token sequence $X=(\textsc{[cls]}, x_1, ..., x_N, \textsc{[sep]}, \textsc{[mode]}, \textsc{[sep]})$ is passed through a BART encoder, where the sub-word tokens of the text are denoted as $x_1, ..., x_N$. \textsc{[cls]} and \textsc{[sep]} are start-of-text and separator pseudo-tokens respectively, while \textsc{[mode]} is one of \textsc{[sys]}/\textsc{[user]}/\textsc{[klg]} to indicate whether the text belongs to a system utterance, user utterance, or factoid respectively. The state of the \textsc{[cls]} is used as the utterance's / factoid's semantic embedding. 
Similarly, to generate the topic embeddings $t_u$ and $t_k$ of every utterances and factoid, the BART encoder sequence input is $T=(\textsc{[cls]}, x_1, ..., x_N, \textsc{[sep]}, \textsc{[mode]}, \textsc{[sep]}, \textsc{[posit]},$ $ \textsc{[sep]})$, where \textsc{[posit]} is the position of the corresponding  dialogue history utterance (zero if the text belongs to a factoid). The state of the \textsc{[cls]} is used as the topic embedding.

\subsection{Topic-aware Attention}

In the next step, TARG calculates feature interactions over the semantic embeddings to extract matching features, which are subsequently weighted by a number of topic-aware attention mechanisms.
These attention mechanisms operate over the topic embeddings of utterances and factoids to calculate topic-aware utterance / factoid pair matching representations. 
The motivation is to incorporate a more flexible way to weight and aggregate matching features of different dialogue history utterances with topic-aware attention, so that the model learns to better attend over them.

Specifically, we design three different types of topic-aware attention that are calculated between each topic embedding $t_k^i$, corresponding to the $i$-th factoid, and the topic embeddings of all utterances in dialogue history $T_u$, as follows:

\hfill

\noindent\textbf{Dot Product}. We concatenate the utterance topic embeddings $t_u^j \in \mathbb{R}^{H}$ with the factoid topic embedding, and compute the dot product between parameter $w_d \in \mathbb{R}^{2H}$ and the resulting vector:
\begin{equation}
A_{d}^{i} = \text{softmax}(\text{exp}([t_u^j,t_k^i]w_d), \forall t_u^j \in T_u )
\end{equation}


\hfill

\noindent\textbf{Bilinear}. We compute the bilinear interaction between $t_u^j$ and $t_k^i$ and then normalize the result:
\begin{equation}
A_{b}^{i} = \text{softmax}(\text{exp}(t_u^j W_b t_k^{i\top}), \forall t_u^j \in T_u )
\end{equation}
where $W_b \in \mathbb{R}^{H \times H}$ is a bilinear interaction matrix.

\hfill

\noindent\textbf{Outer Product}. We compute the outer product between $t_u^j$ and $t_k^i$, then project this feature vector 
through a fully connected layer and a softmax:
\begin{equation}
A_{o}^{i} = \text{softmax}(\text{exp}((t_u^j \times t_k^i)w_o), \forall t_u^j \in T_u )
\end{equation}

\noindent where $w_o \in \mathbb{R}^{H}$ is a parameter and $\times$ is the outer product.


In parallel, we calculate the feature interaction matrix $F_i \in \mathbb{R}^{N \times H}$ between the semantic embeddings of all utterances $s_u^j$ and the factoid $s_k^i$. $N$ is the number of dialogue utterances. Every row $F_{i,j}$ of $F_i$ is calculated as follows:
\begin{gather}
F_{i,j} = v_f^{\top} \text{tanh}(s_u^{j} W_f s_k^{i\top} + b_f)
\end{gather}

\noindent with $W_f \in \mathbb{R}^{H \times H}$, $b_f \in \mathbb{R}$, $v_f\in \mathbb{R}^H$ being model parameters.

To obtain a unified utterance / factoid pair representation $k_i$ for each factoid $i$, we concatenate the weighted sums of all utterances / factoid interaction embeddings with the different attention mechanisms.
The final topic-aware utterance / factoid pair representation across all factoids is $K \in \mathbb{R}^{3H\times M}$, where $M$ is the number of factoids. The $i$-th column vector $k_i$ is calculated as follows:
\begin{gather}
k_i = [A_d^{i\top} F_i, A_b^{i\top} F_i, A_o^{i\top} F_i]
\end{gather}

\subsection{Relevant Knowledge Selection}

For the purpose of knowledge selection, TARG treats all external knowledge as a single document, by simply concatenating all available factoids. To account for the possibility that the system response shouldn't be grounded on any external knowledge, a ``non-relevant'' pseudo-factoid is included.

The relevant knowledge selector takes the topic-aware representations of these sequential factoids as input and predicts a span over the overall document that the system response should be grounded on. Through this process, several knowledge candidates may appear in the selected span.

The grounded span is derived by predicting the start and the end indices of the span in the document. We obtain the probability distribution of the start index and end index over the entire document by the following equations:
\begin{gather}
p^s = \text{softmax}(W_s^{\top} K + b_s^{\top} ),\\
p^e = \text{softmax}(W_e^{\top} K + b_e^{\top} ),
\end{gather}

\noindent where $W_s, W_e \in \mathbb{R}^{3H}$, $b_s, b_e \in \mathbb{R}^{M}$ are trainable weight vectors.

\subsection{System Response Generation}

The system response generator decodes the response by attending on the selected knowledge span. Since the span may contain several factoids, we first use a Convolution Neural Network (CNN) to fuse the information. We apply this CNN even when only a single factoid is present in the span for consistency. The CNN receives the topic-aware utterance / factoid pair embeddings of the selected span, and outputs the fusion embedding $f \in \mathbb{R}^{H} $:
\begin{equation}
f = \text{CNN}(K_{:,s:e}),
\end{equation}

\noindent where $s$ and $e$ are the start and end indexes.

We employ a BART decoder for the system response generator, which takes the fusion embedding $f$ as its initial hidden state. At each decoding step $t$, the decoder receives the embedding of the previous item $w_{t-1} \in \mathbb{R}^{H}$, the previous hidden state $h_{t-1} \in \mathbb{R}^{H}$, and the topic-aware utterance / factoid pair embeddings of the selected span $K_{s:e,:}$, and produces the current hidden state $h_t \in \mathbb{R}^{H}$:
\begin{equation}
h_t = \text{BART}(w_{t-1}, h_{t-1}, K_{:,s:e}).
\end{equation}

A linear transformation layer produces the generated word distribution $p_v$ over the vocabulary:
\begin{equation}
p_v = \text{softmax}(V W_v h_t + b_v),
\end{equation}
where $V \in \mathbb{R}^{L\times H}$ is the word embeddings of the vocabulary, $L$ is the vocabulary size, and $W_v \in \mathbb{R}^{H\times H}$ and $b_v \in \mathbb{R}$ are transformation parameters.

\subsection{Optimization}
For each turn, our model selects the relevant knowledge and generates the current turn's response. We optimize the knowledge selector and response generator via their cross-entropy losses $\mathcal L_s$, $\mathcal L_g$:
\begin{gather}
\mathcal L_s = -\frac{1}{NM}\sum_{n=0}^{N}\sum_{m=0}^{M}[\text{log}(p^s_{y^s_{nm}}) + \text{log}(p^e_{y^e_{nm}}) ], \\
\mathcal L_g = -\frac{1}{NM}\sum_{n=0}^{N}\sum_{m=0}^{M}\text{log}P(Y_{nm}|D_{nm},K_{nm}),
\end{gather}
where $N$ is the number of samples, $M$ is the number of dialogue turns, $y^s_{nm} / y^e_{nm}$ and $p^s / p^e$ respectively represent the ground truth and predicted start/end positions at $m$-th dialogue turn of sample $n$, $D_{nm}$ is the input dialogue context, $K_{nm}$ is the input knowledge, and $Y_{nm}$ is the ground truth system response at $m$-th dialogue turn of sample $n$.
We compute the joint loss $\mathcal L$ as follows:
\begin{equation}
\mathcal L = \lambda \cdot \mathcal L_s + (1-\lambda) \cdot\mathcal L_g,
\end{equation}

\noindent where $\lambda \in [0, 1]$ is a balance coefficient.


\begin{table}[!t]
\centering
\resizebox{0.48\textwidth}{!}{
\begin{tabular}{ll|c|cccc}
        \toprule
        \multirow{2}{*}{\bf{Domain}}&\multirow{2}{*}{\bf{\#Dials}}&\multirow{2}{*}{\bf{\#Docs}}&\multicolumn{4}{c}{{ \bf{avg \# per doc}}}\\
        &&&{\bf{tk}} &{\bf{sp}}&{\bf{p}}&{\bf{sec}}\\
 		\hline
        \hline
        \text{ssa} & 1192&  109& 795 & 70 & 17 & 5 \\
        \text{va} & 1330&  138& 818 & 70 & 20 & 9\\
        \text{dmv} & 1305&  149& 944 & 77 & 18 & 10\\
        \text{studentaid} & 966 & 91&  1007 & 75 & 20 & 9 \\
        \hline
        \text{all} & 4793&  487& 888 & 73 & 18 & 8\\
        \toprule
	\end{tabular}
}
\caption{Number of dialogues, documents and average of content elements per document (tk: tokens, sp: spans, p: paragraphs, sec: sections) per domain in Doc2Dial.}
\label{tab:doc2dial}
\end{table}

\begin{table}[!t]
\centering
\resizebox{0.48\textwidth}{!}{
\begin{tabular}{ll|c|cccc}
        \toprule
        \multirow{2}{*}{\bf{Domain}}&\multirow{2}{*}{\bf{\#Dials}}&\multirow{2}{*}{\bf{\#Snippets}}&\multicolumn{2}{c}{{ \bf{\#per-snip}}}\\
        &&&{\bf{tk}} &{\bf{sent}}\\
 		\hline
        \hline
        \text{Hotel} & -&  1219&  9& 1.00\\
        \text{Restaurant} & -&  1650& 7 & 1.00\\
        \text{Train} & -&  26& 15 & 1.20\\
        \text{Taxi} &  -& 5&   19 & 1.15\\
        \hline
        \text{all} & 10,438 &   2900&  8 & 1.00\\
        \toprule
	\end{tabular}
}
\caption{Number of dialogues, snippets and average number of content elements per snippet (tk: tokens, sent: sentences) per domain in the DSTC9 dataset.}
\label{tab:dstc9}
\end{table}

f

\section{Experiments}

\subsection{Datasets}

We evaluate our proposed approach on two benchmark data sets on task-oriented dialogue: Doc2Dial~\citep{feng2020doc2dial} and DSTC9~\citep{gunasekara2020overview}. Doc2Dial is a leaderboard dataset with a withheld test set used for ranking participating systems, which includes conversation dialogues between an assisting system and an end user, with an accompanying set of documents wherein distinct factoids are clearly annotated; further annotations indicate which dialogue utterances are grounded on which factoids of the associated documents. The Doc2Dial dataset includes many cases of conversations that are grounded on factoids from different documents. By considering the title of each document as a distinct topic, each of these conversations can be interpreted to involve many interconnected topics under a general inquiry, making it an ideal dataset for our approach. 

The DSTC9 dataset also includes conversation dialogues, but the external knowledge is in the form of FAQ documents, in essence containing question answering pairs on a specific domain; we consider each pair as a distinct factoid and their domain as the topic. In practice, these FAQs are to be used to answer follow-up user questions that are out of the coverage of a dialogue system's database. Similarly to Doc2Dial, the ``topic'' in the DSTC9 dataset is also varied thoughout the conversations. 

As mentioned before, we interpret the title of the factoid's originating document or its annotated domain as the topic of the factoid. However, this assumption would be reasonable only if the factoids are relatively short. Table~\ref{tab:doc2dial} and Table~\ref{tab:dstc9} presents the statistics of the Doc2Dial and DSTC9 datasets, and we can observe that on average the knowledge factoids are indeed relatively short in both datasets. 

Information on the evaluation measures and implementation details can be found in the Appendix.

\subsection{Baselines}
In the following experiments, we compare our approach against previously published state-of-the-art approaches on the Doc2Dial and DSTC9 datasets. We have not re-implemented these approaches, but report their already published results for the datasets for which they are available.\footnote{While there are better performing systems in the DSTC9 and Doc2Dial leaderboards, these are either not published, not based on a single method, or exploit additional external data, and thus are not directly comparable to this work.}

\hfill 

\noindent\textbf{Base-D2D}~\citep{feng2020doc2dial}: This is the baseline provided by the Doc2Dial challenge. It consists of an extractive question answering model using a BART~\cite{devlin-etal-2019-bert} encoder to predict the grounding span in the document and a BART model to generate system responses. Base-D2D-ST directly uses the topic of the previous turn as the topic of current turn. 

\noindent\textbf{JARS}~\citep{khosla2021team}: A transformer-based~\citep{lan2019albert} extractive question-answering model that extracts relevant spans from the documents. They focus on knowledge selection and do not perform response generation.

\noindent\textbf{CAiRE}~\citep{xu2021caire}: An ensemble approach of fine-tuned RoBERTa~\citep{liu2019roberta} models, trained with a meta-learning objective over data-augmented datasets.

\noindent\textbf{RWTH}~\citep{daheim2021cascaded}: They use a biaffine classifier to model spans, followed by an ensemble for knowledge selection, and a cascaded model that grounds the response prediction on the predicted span for response generation.

\noindent\textbf{Base-DSTC}~\citep{gunasekara2020overview}: The baseline provided by the DSTC9 challenge is a response generation model obtained by fine-tuning the GPT-2~\citep{budzianowski2019hello} model with a standard language modeling objective. Base-DSTC-ST directly uses the topic of the previous turn as the topic of current turn. 

\noindent\textbf{KDEAK}~\citep{chaudhary2021unstructured}: A model which formulates knowledge selection as a factorized retrieval problem with three modules performing domain, entity and knowledge level analyses. The response is generated using a GPT-2 model attending on any relevant retrieved knowledge.

\noindent\textbf{RADGE}~\citep{Radge}: A multi-task method that exploits correlations between dialogue history and keywords extracted from the API through fine-tuning a sequence of ELECTRA models~\cite{clark2020electra}. 

\noindent\textbf{EGR}~\citep{egr}: An approach that uses relevance similarity to score factoids, and later reranks them with a rule-based algorithm based on entity names parsed from the dialogue. The response is generated with a BART model.

\begin{table}[!t]
\centering
\resizebox{0.48\textwidth}{!}{
\begin{tabular}{l|cc|c}
        \toprule
        \multirow{2}{*}{\bf{}}&\multicolumn{2}{c|}{{ \bf{Knowledge}}}&\multicolumn{1}{c}{{ \bf{Response}}}\\
        \multirow{2}{*}{\bf{Model}}&\multicolumn{2}{c|}{{ \bf{Selection}}}&\multicolumn{1}{c}{{ \bf{Generation}}}\\
        &{EM} &{F1}&{BLEU-4}\\
 		\hline
        \hline
        \text{Base-D2D} & 37.2&  52.9 & 17.7\\
        \text{Base-D2D-ST} & 27.6&  35.2 & 12.1 \\
        \text{JARS} & 42.1&  57.8 & -\\
        \text{CAiRE} & 45.7&  60.1 & 22.3\\
        \text{RWTH} & 46.6&  62.8& 24.4\\
        \hline
        \text{\bf TARG} & ${\bf 49.8}$ & ${\bf 66.4}$& ${\bf 28.6}$\\
        \toprule
	\end{tabular}
}
\caption{Performance of TARG and related work on Doc2Dial. \textbf{Bold} denotes best results in that metric.}
\label{tab:results_doc2dial}
\end{table}

\begin{table*}[!h]
\centering
\resizebox{1.0\textwidth}{!}{
\begin{tabular}{l|cc|ccccccc}
        \toprule
        \multirow{2}{*}{\bf{}}&\multicolumn{2}{c|}{{ \bf{Knowledge}}}&\multicolumn{7}{c}{{ \bf{Response}}}\\
        \multirow{2}{*}{\bf{Model}}&\multicolumn{2}{c|}{{ \bf{Selection}}}&\multicolumn{7}{c}{{ \bf{Generation}}}\\
        &{MRR@5} &{Recall@5}&{BLEU-1}&{BLEU-2}&{BLEU-3}&{BLEU-4}&{ROUGE-1}&{ROUGE-2}&{ROUGE-L} \\
 		\hline
        \hline
        \text{Base-DSTC} & 0.726&  0.877& 0.303 & 0.173 & 0.100 & 0.065 & 0.338 & 0.136 & 0.303\\
        \text{Base-DSTC-ST} & 0.612&  0.743& 0.251 & 0.132 & 0.083 & 0.047 & 0.262 & 0.104 & 0.244\\
        \text{KDEAK} & 0.853&  0.896& 0.355 & {\bf 0.230} &  0.153 & 0.104 &  0.397 & {\bf0.190} &0.357\\
        \text{RADGE} & {\bf 0.937}&  0.966& 0.350 & 0.217 & 0.135 & 0.089 & 0.393 &0.175&0.355\\
        \text{EGR} & 0.894&  0.934& 0.361 & 0.226 & 0.140 & 0.096 & 0.397&0.179&0.353\\
        \hline
        \text{\bf TARG} & $0.935$ & ${\bf 0.972}$& ${\bf 0.366}$ & $0.224$ & ${\bf 0.156}$ & ${\bf 0.111}$ &{\bf 0.408} &{0.183} &{\bf 0.360}\\
		\toprule
	\end{tabular}
}
\caption{Performance of TARG and related work on the DSTC9 dataset. \textbf{Bold} denotes best results in that metric.}
\label{tab:results_datc9}
\end{table*}

\subsection{Experimental Results}
Tables~\ref{tab:results_doc2dial} and ~\ref{tab:results_datc9} show our results on Doc2Dial and DSTC9 respectively. Observe that TARG performs significantly better than related work in both knowledge selection and response generation on the Doc2Dial dataset, outperforming the second best system by 3.2, 3.6, and 4.2 points in EM, F1 and BLEU-4 respectively. 

On the DSTC9 dataset, TARG outperforms the related work in most metrics, though by narrow margins. Due to the smaller differences, we consider TARG to be performing on par with state-of-the-art on DSTC9.
The performance gains of TARG can be explained by the topic-aware mechanism as it provides a more flexible way to weight and aggregate different dialogue history turns. This indicates that better understanding of the dialogue history is crucial for predicting the relevant factoids and generating a reasonable response. 

The main difference between datasets is the frequency of topic shifts. The average number of topics per dialogue is 8.83 and 2.58 on Doc2Dial and DSTC9 respectively. This difference can be partially explained by how we infer each dataset's topic, e.g.  since the topic in DSTC9 is the domain of each question-answer pair, and multiple pairs belong to the same domain, the topic shifts are considerably more limited than in the Doc2Dial dataset. 
We further examined how BLEU scores are effected if we isolate DSTC9 dialogues that have more than the average number of topics. Specifically, we evaluated TARG on DSTC9 dialogues which exclusively have 2, 3, and 4 topics, and the BLEU is 0.363, 0.372, and 0.378 respectively. This indicates that more topic shifts provide more signal for the model to exploit.

An additional difference between the datasets is that the topic for each factoid in Doc2Dial can be considered to be fine-grained, 
e.g. ``VA clothing allowance'', ``About your eligibility'', and ``How to get these benefits'', while in the DSTC9 dataset, the topic for each factoid can be considered coarse-grained, e.g.
``Restaurant'', ``Hotel'', ``Taxi'', and ``Train''. 
These differences collectively show that the lower performance on DSTC9 is due to its coarse-grained topics, and the lower number of average topic shifts. This suggests that a further division of the documents on more fine-grained topics and introducing more topic shifts in DSTC9 dialogs would help TARG perform better. However, we cannot straightforwardly examine how these two improvements interact with each other, and leave such analysis for future work.

\section{Discussion}

\begin{figure}[!t]
\centering
\includegraphics[width=0.48\textwidth]{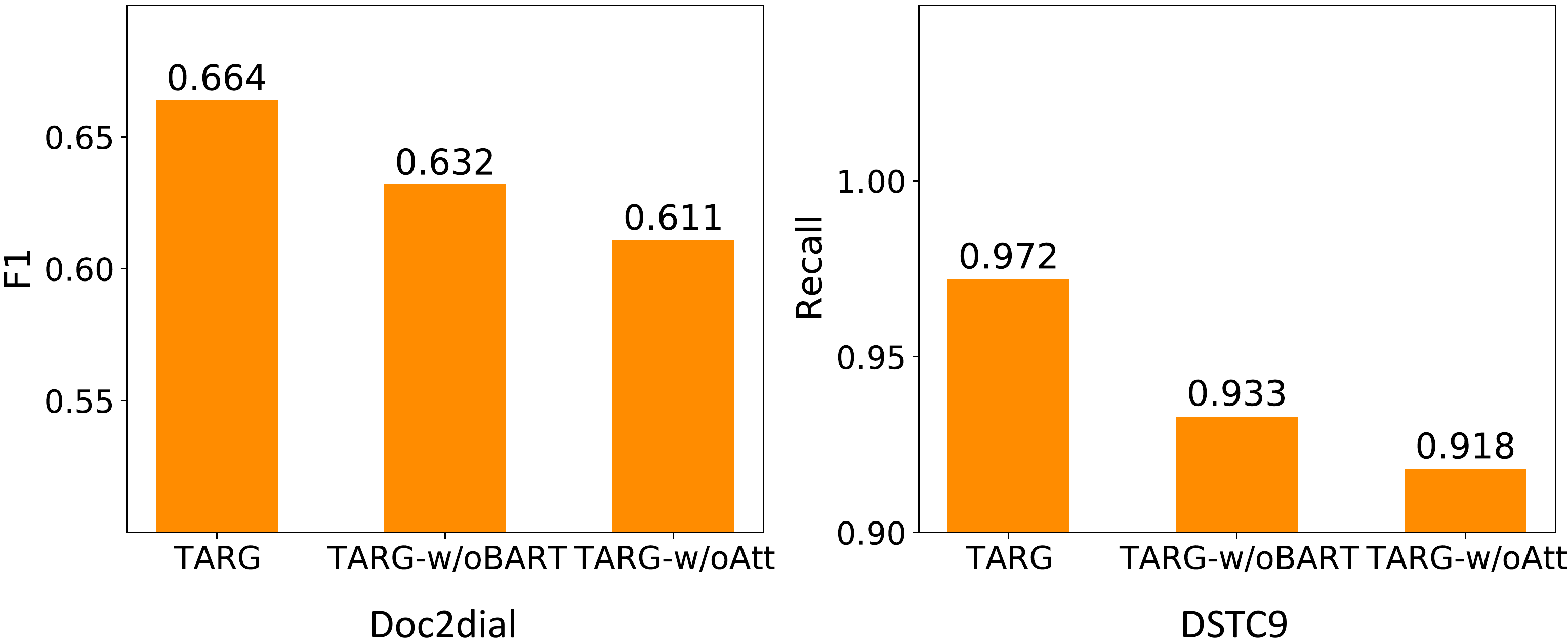}
\caption{Ablation study for knowledge selection.}
\label{fig:abla_ks}
\end{figure}

\begin{figure}[!t]
\centering
\includegraphics[width=0.48\textwidth]{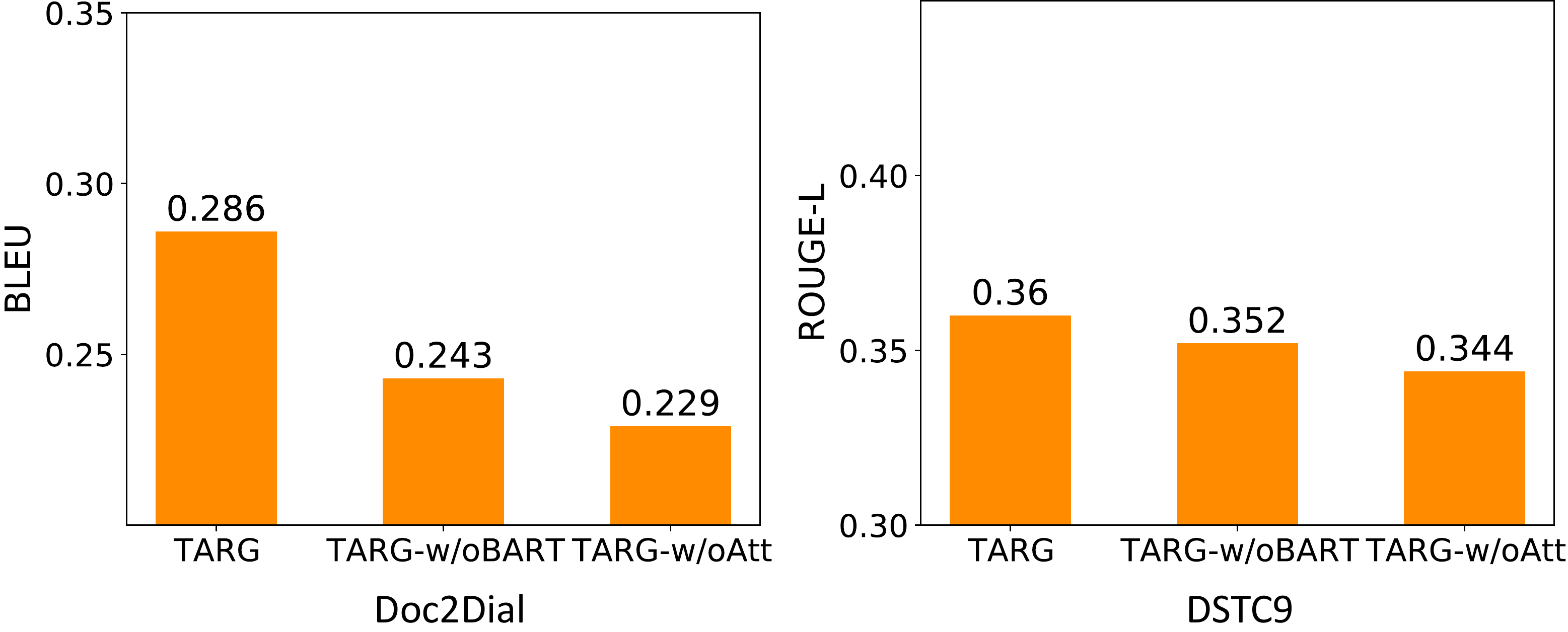}
\caption{Ablation study for response generation.}
\label{fig:abla_g}
\end{figure}

\begin{figure*}[!th]
\centering
\includegraphics[width=0.98\textwidth]{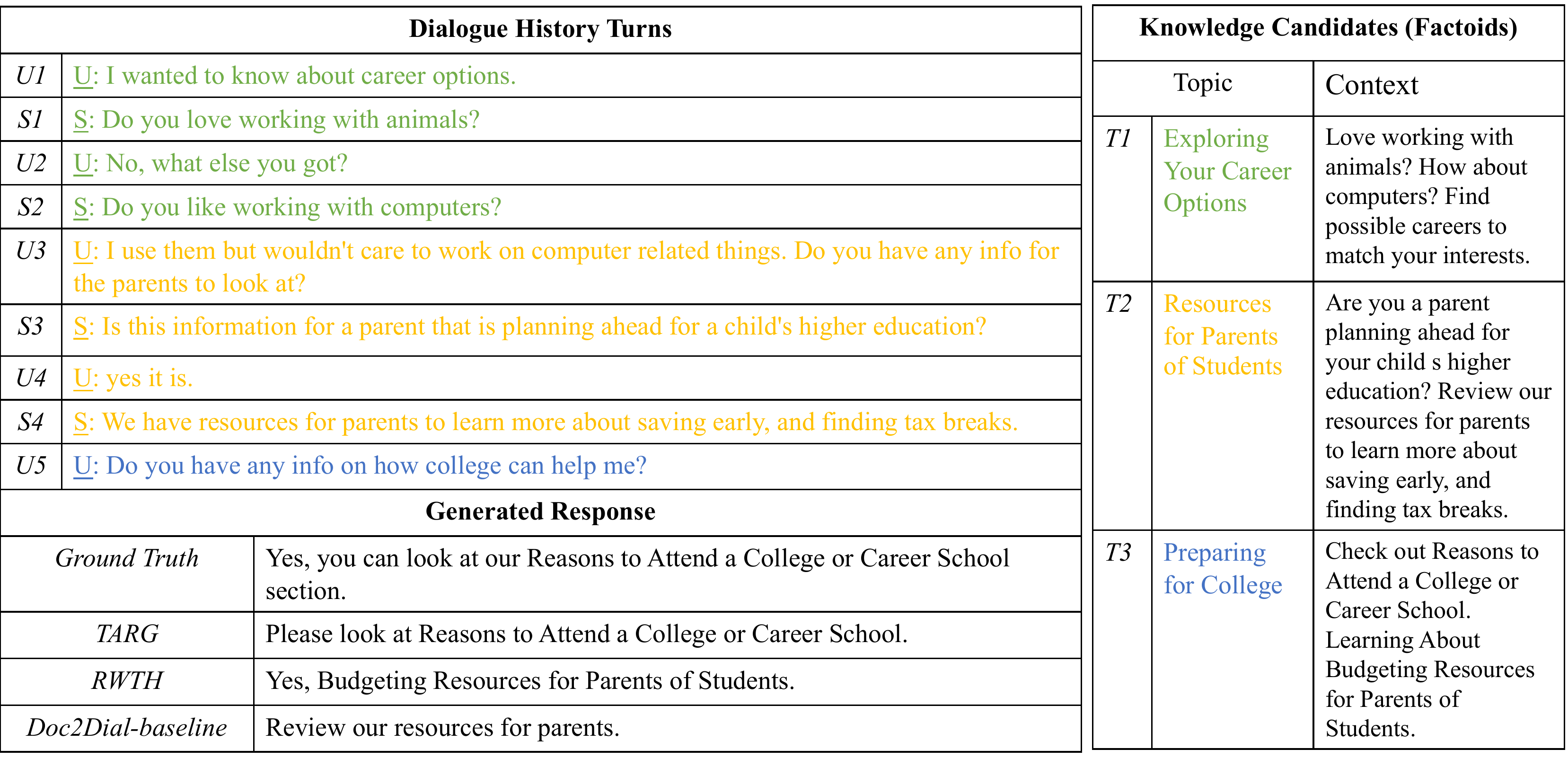}
\caption{Case study on Doc2Dial. Dialogue history turns are grounded to knowledge candidates of the same color.}
\label{fig:case}
\end{figure*}

\begin{figure}[!th]
\centering
\includegraphics[width=0.48\textwidth]{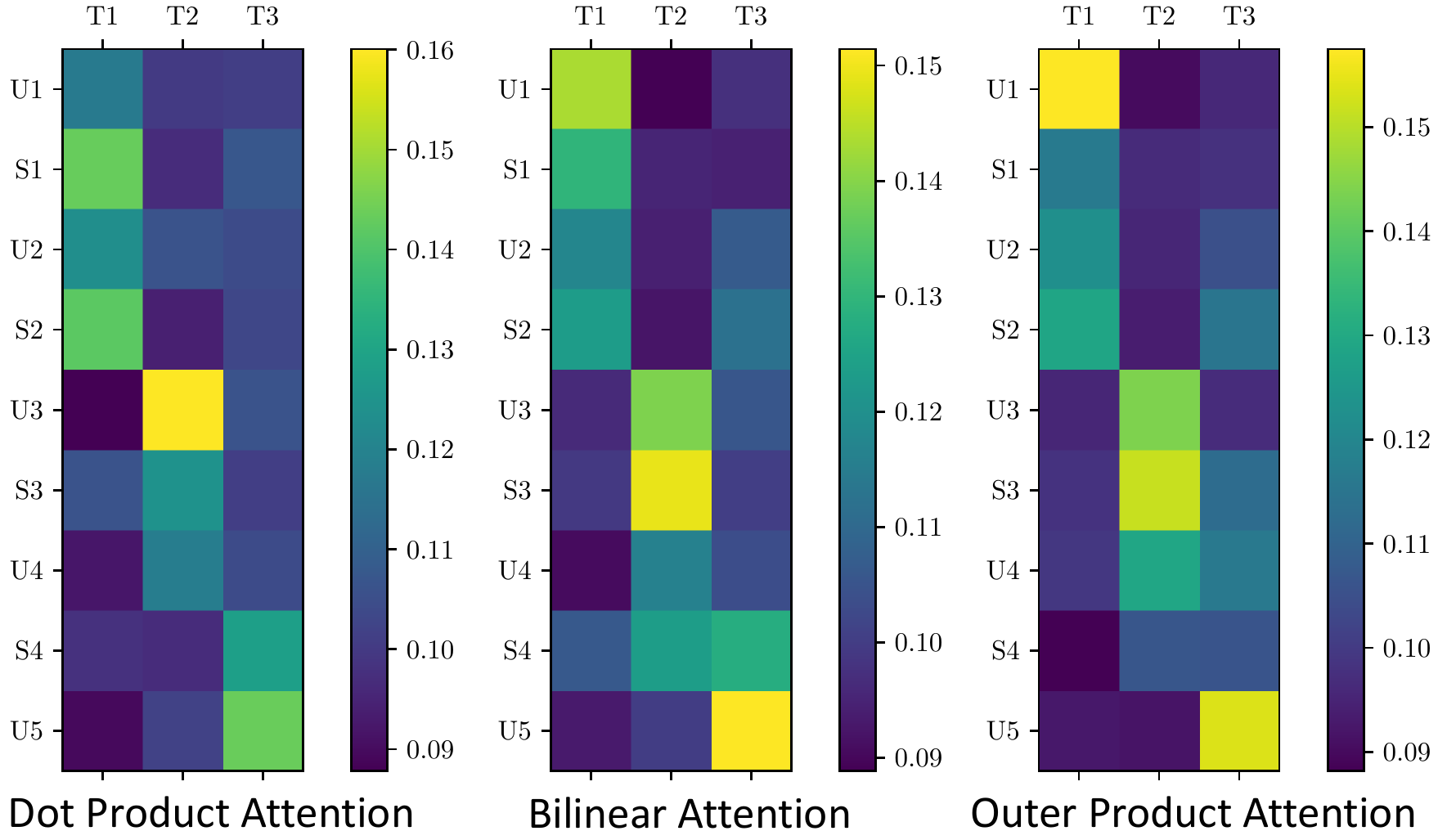}
\caption{Visualization of learned topic-aware attention of dialogue history utterances U-X and S-X (for user and system utterance) for each topic T-X in the example in Figure~\ref{fig:case}. 
Lighter spots mean higher attention scores.}
\label{fig:visualization}
\end{figure}

\subsection{Ablation Study}
Here we conduct an ablation study of TARG, to explore the effects of the BART model, topic-aware attention, as well as the different topic attention mechanisms. The results indicate that all these mechanisms are necessary to the performance of knowledge selection and response generation.

\textit{Effect of BART}: 
To investigate the effectiveness of using BART in the utterance / factoid encoder and system response generator, we replace BART with a bi-directional LSTM and rerun the model for Doc2Dial and DSTC9. As shown in Figures~\ref{fig:abla_ks} and ~\ref{fig:abla_g}, the performance of the BiLSTM-based model TARG-w/oBART decreases significantly in knowledge selection, and especially in response generation as is indicated by the drop in BLEU. As expected, this indicates that the BART model can create and utilize more accurate representations for dialogue history and unstructured knowledge.

\textit{Effect of topic-aware attention}: 
Next we remove the topic-aware attention mechanisms (TARG-w/oAtt). Figures~\ref{fig:abla_ks} and ~\ref{fig:abla_g} again show that the respective performances deteriorate considerably. This shows that topic-aware attention helps derive an important weighting scheme over the utterances leading to better understanding of dialogue history.

\begin{table}[!t]
\centering
\resizebox{0.48\textwidth}{!}{
\begin{tabular}{l|cc|c}
        \toprule
        \multirow{2}{*}{\bf{}}&\multicolumn{2}{c|}{{ \bf{Knowledge}}}&\multicolumn{1}{c}{{ \bf{Response}}}\\
        \multirow{2}{*}{\bf{Model}}&\multicolumn{2}{c|}{{ \bf{Selection}}}&\multicolumn{1}{c}{{ \bf{Generation}}}\\
        &{EM} &{F1}&{BLEU}\\
 		\hline
        \hline
        \text{TARG-dot} & 0.468&  0.642 & 0.261\\
        \text{TARG-bilinear} & 0.481&  0.652 & 0.268\\
        \text{TARG-outer} & 0.489&  0.655 & 0.275\\
        \hline
        \text{\bf TARG} & ${\bf 0.498}$ & ${\bf 0.664}$& ${\bf 0.286}$\\
        \toprule
	\end{tabular}
}
\caption{Ablation over different attention mechanisms.}
\label{tab:results_attention}
\end{table}


\textit{Effect of topic attention mechanisms}: 
Here we compare TARG against TARG-dot, TARG-bilinear, and TARG-outer which use exclusively doc product attention, bilinear attention, and outer product attention respectively. Table~\ref{tab:results_attention} shows 
that dot product attention underperforms compared to bilinear and outer product attention while bilinear attention's performance is comparable with outer product attention. In addition, any isolated attention mechanism performs considerably worse than their fusion, supporting its utilization. We conjecture that this is due to how different attention mechanisms focus on different topic features.

\subsection{Analysis on Topic Shift}
To facilitate a better understanding of how topic shifts occur in our model, we present a case study from the Doc2Dial dataset. On the top of Figure~\ref{fig:case} are the previous turns of dialogue history, while on the right is a subset of the available factoids. We can observe how the topic changes throughout the turns of dialogue history (by consulting the corresponding factoid topic), from ``Exploring Your Career Options'' in turns 1 and 2, to ``Resources for Parents of Students'' in turns 3 and 4, and finally ``Preparing for College'' in turn 5. 

On the bottom of Figure~\ref{fig:case}, we present responses generated by our proposed model TARG, the best of the previous work RWTH, the Doc2Dial-baseline, and the ground truth. 
Observing the responses and comparing with the ground truth, Doc2Dial-baseline seems to generate irrelevant response, picking the wrong topics from the candidates on the right, i.e. ``Resources for Parents of Students''. RWTH picks right topic, but it selects wrong factoid ``Review our resources for parents'' to generate response. 
TARG generates the more relevant and fluent response of the three, as its topic-aware attention informs knowledge selection to pick the topic and factoid that more naturally follows the dialogue history, i.e. ``Reasons to Attend a College or Career School''. Furthermore, TARG's BART decoder ensures the fluency of the output.

Figure~\ref{fig:visualization} presents a visualization of TARG's learned topic-aware attention over the dialogue utterances and topics of the case study. This includes Dot Product Attention, Bilinear Attention, and Outer Product Attention. We can see that topic-aware attention captures reasonable dialogue utterance weights for each topic, with the weighing moving from topic T1 to T2 and to T3 as attentions are calculated over the dialogue history utterances. This supports our claim that modeling the topic shifts can be helpful for knowledge selection, and consequently response generation, through better understanding of the dialogue history.

\section{Conclusion}
In this paper, we proposed TARG: ``Topic-Aware Response Generation'', a topic-aware model which incorporates multiple topic-aware attention mechanisms to derive the importance weighting scheme over both dialogue utterances and unstructured external knowledge, and through that facilitate better dialogue history understanding. Our proposed method achieves state-of-the-art results in both knowledge selection and response generation, outperforming previous state-of-the-art by 3.2, 3.6, and 4.2 points in EM, F1 and BLEU-4 respectively on Doc2Dial, and performing comparably with previous work on DSTC9. To provide further insights, we also presented an ablation study of our model that supported the importance of our method's various components, and discussed a case study accompanied by an analysis of the attention mechanisms.

\section*{Limitations}
The main limitation of the proposed method is its reliance on annotated or easily inferrable topics in the external knowledge sources. Future work should explore how this method can be applied when such topics are absent, e.g. by inferring topics through Latent Dirichlet Analysis. Our analysis also shows that our method performs better when these topics are fine-grained and a large number of topic shifts are expected in the dialogue.
A more technical limitation of our model is that due to the limited input context size of the pre-trained language model we used, its scalability to long dialogue context is difficult.
Finally, due to data availability, we only conducted experiments on English dialogues. While little in our method should be affected by the limited morphology of the English language, our results should be confirmed to hold on more structurally complicated languages.

\section*{Acknowledgements}
The authors would like to thank the reviewers for their suggestions on how to improve the paper. They would also like to thank the MindSpore team for providing technical support\footnote{\url{https://www.mindspore.cn/en}}\footnote{\url{https://github.com/mindspore-ai}}.

\bibliography{anthology}
\bibliographystyle{acl_natbib}

\appendix
\clearpage
\section{Implementation Details}
We use a pre-trained BART-base model 
to encode utterances and factoids. The max sentence length is set to 50 and the max number of dialogue turns is set to 15. 
The hidden size of attentions are all set to 768. 
The size of the convolution and pooling kernels are set to (3, 3, 3).
The joint loss $\mathcal \lambda$ is 0.5. The dropout probability is 0.1. The batch size is set to 8. We optimize with Adam 
and an initial learning rate of 3e-5.

\section{Evaluation Measures}
We make use of the following automatic evaluation metrics in our experiments. For each dataset, we calculate the metrics used by the respective challenges for consistency.


\noindent\textbf{Exact Match (EM)}: This measures what part of the predicted knowledge span matches the ground truth factoid exactly.

\noindent\textbf{Token-Level F1}: 
We cast the predicted spans and ground truth factoids as bags of tokens, and compute F1 between them. 

\noindent\textbf{MRR@5}: A metric based on the rank of the first ground truth factoid in a system's top-5 ranking.

\noindent\textbf{Recall@5}: This metric counts how many ground truth factoids occur in a system's top-5 ranking.

\noindent\textbf{BLEU-X}~\cite{papineni-etal-2002-bleu}: BLEU-X estimates a generated response's via measuring its n-gram precision against the ground truth. X denotes the maximum size of the considered n-grams (i.e. unigrams, bigrams, trigrams, 4-grams).

\noindent\textbf{ROUGE-X}~\cite{lin-2004-rouge}: ROUGE-X measures n-gram recall between generated and ground truth response. ROUGE-L measures the longest common word subsequence.

\section{Analysis of Knowledge Selection}
\label{sec:appendix}

We further conduct an analysis on how the selected knowledge span differs from turn to turn as this also indicates a shift in topic. Table~\ref{tab:knowledge} shows the average number of knowledge span changes as observed in the grounded truth and in the predicted output of Base-D2D and TARG, on the Doc2Dial dataset. We can see that the knowledge span changes are frequent in the ground truth, 
and that TARG's average knowledge span changes is closer to that of the ground truth. This indicates that TARG can more accurately follow the knowledge span changes in the dataset than Base-D2D.

We further investigate the number of the selected factoids per turn in Doc2Dial, i.e. the average number of factoids covered by the predicated spans. As shown in Table~\ref{tab:knowledge}, we can again see that TARG's behavior is closer to that of the ground truth.

\begin{table}[]
\centering
\resizebox{0.48\textwidth}{!}{
\begin{tabular}{l|c|c}
        \toprule
        {\bf Model} & {\bf Knowledge Changes} & {\bf Factoid}\\
 		\hline
        \hline
        \text{Ground Truth} & 9.22 &   1.46\\
        \text{Base-D2D} & 8.73 &   1.23\\
        \text{TARG} & 9.02&  1.54\\
        \toprule
	\end{tabular}
}
\caption{Average number of knowledge changes per dialogue and average number of factoid per turn in Doc2Dial.}
\label{tab:knowledge}
\end{table}

\end{document}